\documentclass{article} 
\usepackage[preprint]{colm2026_conference}

\usepackage{microtype}
\usepackage{hyperref}
\usepackage{url}
\usepackage{booktabs}

\usepackage{lineno}

\usepackage{hyperref}       
\usepackage{url}            
\usepackage{multirow}
\usepackage{graphicx}
\usepackage{tcolorbox}
\usepackage{subcaption}
\usepackage{amsmath}

\usepackage{xcolor}

\definecolor{darkblue}{rgb}{0, 0, 0.5}
\hypersetup{colorlinks=true, citecolor=darkblue, linkcolor=darkblue, urlcolor=darkblue}

\title{Back to the Barn with LLAMAs: Evolving Pretrained LLM Backbones in Finetuning Vision‑Language Models}

\author{
Sameera Horawalavithana*, Lauren Phillips, Ian Stewart, Sai Munikoti, \& Karl Pazdernik \\
Pacific Northwest National Laboratory \\
Richland, WA 99352, USA \\
\texttt{\{yasanka.horawalavithana, lauren.phillips, ian.stewart,} \\
\texttt{sai.munikoti, karl.pazdernik\}@pnnl.gov}
}

\begin{document}

\ifcolmsubmission
\linenumbers
\fi

\maketitle

\begin{abstract}
Vision-Language Models (VLMs) have rapidly advanced by leveraging powerful pre-trained Large Language Models (LLMs) as core reasoning backbones. As new and more capable LLMs emerge with improved reasoning, instruction-following, and generalization, there is a pressing need to efficiently update existing VLMs to incorporate these advancements. However, the integration of new LLMs into VLMs, particularly how the evolving LLMs contribute to multimodal reasoning, alignment, and task-specific performance remains underexplored. 
Addressing this gap is important for VLM development, given the rapid evolution of pretrained LLM backbones.
This study presents a controlled and systematic investigation of how changes in the pretrained LLM backbone affect downstream VLM task performance.
By having the vision encoder, training data, and post-training algorithm remain same across LLAMA-1, LLAMA-2, and LLAMA-3 based VLMs, we find that newer LLM backbones do not always lead to better VLMs, but the performance depends on the downstream VLM task.
For example, in visual question and answering tasks, newer LLM backbones tend to solve different questions rather than just more questions, and our analysis shows this is driven by differences in how the models process information, including better calibrated confidence and more stable internal representations.
We also find that some VLM capabilities appear only in the newest LLM generation, while tasks that depend mainly on visual understanding see little benefit from a newer LLM backbone.
\end{abstract}

\section{Introduction}
Modern VLMs build upon the strong reasoning and language understanding capabilities of LLMs, extending them to interpret and process visual information through large-scale pre-training, across applications such as image captioning, visual question answering, object detection, and multimodal content generation~\citep{Qinetal2025,lin2024vila,Hudsonetal2019}.
The success of LLMs has inspired recent VLMs to regard powerful LLMs as their core backbone, utilizing large-scale image-text pairwise datasets for pre-training and applying instruction tuning with large image-text instruction datasets to generate satisfactory answers~\citep{Chengetal2023,Jiaetal2021,Changpinyoetal2021,Ouyangetal2022,Alayracetal2022}.
This two-stage training process supports recent VLMs to achieve impressive performance on downstream vision-language tasks~\cite{Chengetal2023}.

For VLMs to solve a general visual question, two essential capabilities are required: 1) Perception: extracting necessary information from the image; 2) Reasoning: generating answers based on the extracted information and contextual understanding~\citep{Qiaoetal2024}.
An important finding in VLM research is that the performance of individual unimodal pre-trained backbones directly correlates with the performance of the resulting multimodal model~\citep{Wangetal2025,Qinetal2025,Pingetal2025,Chenetal2025}.
For example, when researchers replace language model components: upgrading from LLAMA-1-7B (35.1\% on MMLU) to Mistral-7B (60.1\% on MMLU) leads to substantial improvements across some benchmarks~\citep{Inoueetal2024,Yietal2024}.
Similarly, replacing vision encoders from CLIP-ViT-H (78.0\% on ImageNet) with SigLIP-SO400M (83.2\% on ImageNet) produces significant performance gains across all benchmarks without changing the total parameter count~\citep{Inoueetal2024}.
Cambrian-1~\citep{tong2024cambrian} evaluate LLM backbones of various scales: LLAMA-3-Instruct-8B,
Vicuna-1.5-13B, and Hermes-2-Yi-34 and different vision encoders
such as CLIP, SigLIP, DINOv2, which specifically call for the need for advancements in vision-only models
to ensure better visual understanding.


Existing research comparing LLM impact on VLM performance suffers from significant confounding variables.
Previous studies typically compare models like LLaVA, Qwen-VL, and InternVL that differ not only in their LLM backbone but also in vision encoders, training data, hyperparameters, and implementation details, making it impossible to isolate what actually drives performance differences.
To address this, we conduct a controlled empirical study using a series of LLMs developed by a single entity (e.g., Meta's LLAMA series), identical multimodal datasets, training algorithms (e.g., SFT), and a unified training/evaluation codebase. 
We believe this controlled setup isolates the effect of the evolving LLM backbone on VLM performance better than previous studies, helping us to perform a rigorous analysis of how architectural and pretraining improvements in newer LLMs translate to multimodal capabilities—and where they fall short.
In this study, our primary research question is; \textit{How do systematic improvements in LLMs impact VLM downstream task performance?}.

Our study provides evidence of how LLM evolutions directly impact VLM capabilities, quantifying which abilities (reasoning, instruction following) benefit most and which (low-level perception) show minimal gains.
We focus more on the downstream VLM performance when fine-tuned on a variety of domains such as science, seismic and general scenees.

\section{Related Work}




Recent research has some insights on the relationship between LLM capabilities and VLM performance.
While some research shows having a strong LLM helps in the downstream VLM performance, other research demonstrate negligible performance advantage with newer LLMs.
In this section, we summarize the two sides of research;

\textbf{Systematic Improvements in LLMs improve VLM performance}
\citet{laurencon2024building} demonstrate that replacing LLAMA-1-7B with Mistral-7B as the language backbone leads to substantial performance improvements across benchmarks, establishing a correlation between standalone LLM performance and resulting VLM capabilities. 
This work provides empirical evidence that LLM quality directly impacts VLM performance.
\citet{lin2024vila} present VILA, conducting ablation studies across different LLM backbones including Vicuna-1.5-7B (based on LLAMA-2). 
Their findings on pre-training strategies demonstrate scalability across LLM architectures, suggesting that training improvements are robust to backbone changes.
Upgrading from Qwen-1.5 to Qwen2 improves visual grounding, long-context image understanding, and video tasks.

\textbf{Not significant VLM performance advantage by the LLM improvements}
~\citet{cocchi2025llavamore} introduce LLaVA-MORE, conducting systematic comparisons of different LLM backbones including LLAMA-3.1 in VLM architectures. Their work demonstrates that smaller models with advanced architectures can outperform larger models with older architectures, highlighting the importance of architectural evolution beyond parameter scaling.
\citet{liu2024llavanext} (LLaVA-NeXT) explicitly report that LLAMA-3 improves MMMU/MathVista massively but yields only marginal gains on TextCaps and NoCaps.
~\citet{lu2023mathvista} show that chart/diagram understanding remains limited by resolution and OCR quality rather than language capability.
Recent domain-specific models (Monkey~\cite{li2024monkey}, DocPedia~\cite{feng2023docpedia}, InternVL-Chat-V1.5~\cite{chen2024internvl}) achieve large gains only after high-resolution training and scientific-figure-specific continued pretraining of the vision encoder.


\section{Experimental Setup}
In this section, we outline an experimental setup to evaluate the effects of evolving LLM backbones, such as Meta’s LLAMA series, on the performance of VLMs in downstream tasks.
The goal is to understand how improvements in LLMs impact tasks that combine vision and language understanding.


\subsection{VLM Architecture}
More recently, VLMs follow a consistent model architecture comprising of a visual encoder that extracts visual features, a connector module for modality alignment, and an LLM backbone that serves as the primary reasoning component.
The LLM backbone used for language-based reasoning and text generation, allowing models to achieve effective vision-language grounding.

The VLM architecture consists of three main components. 
First, the LLM backbone serves as the language decoder. We use different versions of an evolving LLM, such as Meta’s LLAMA series (e.g., LLAMA-1, LLAMA-2, LLAMA-3.1), to study the impact of their advancements in architecture, size, or training data.
Second, a pre-trained visual encoder (i.e., \textit{clip-vit-large-patch14}), processes input images and extracts visual features. 
Visual encoder is kept consistent across experiments to focus on the effect of the LLM backbone.
Third, a simple neural network, called a projector, aligns the visual features from the encoder with the text embeddings of the LLM. 
We use the NVIDIA-NeMo codebase~\citep{nemo_framework} and adopt the NeVA model architecture~\citep{nemo_neva_2024} for the experiments.
NeVA leverages the features of the NeMo framework such as model parallelism, activation checkpointing, AMP O2, Flash Attention, and more~\citep{kuchaiev2019nemo}.

\subsection{Finetuning Process}
The training process is divided into two stages to improve the VLM’s performance. 
In the first stage, known as the warm-up or alignment stage, the goal is to align the visual encoder and LLM decoder. 
We use an image captioning dataset, SciCap~\citep{hsu2021scicap}, for this stage. 
In this stage, the model learns to generate descriptive text for technical plots, building a basic understanding of visual-language relationships. 
This approach is inspired by prior research, which shows that such pre-training on captioning data improves zero-shot performance~\citep{mckinzie2024mm1methodsanalysis}.
Specifically, previous research~\citep{mo2023s,pang2022matcha,lee2023pix2struct} suggested that pre-training on similar technical plots improves performance on other domains (e.g., business, medical, remote sensing). 
Training in this stage involves fine-tuning the projector, the LLM and visual encoder. 
We train for a fixed number of epochs (e.g., 1-3) or until performance on a validation set stabilizes.
All the VLM checkpoints record a comparable performance in the captioning tasks after this pre-training stage (see Appendix~\ref{sec:vlm_scicap} for more details).

The second stage focuses on adapting the model to specific vision-language tasks through Supervised Fine-Tuning (SFT).
We selected SFT as the post-training method for this ablation study because it provides most stable, and computationally efficient experimental setup in comparison to the RL-based methods~\citep{chu2025sft}.

We use three domain-specific datasets for evaluation
\begin{itemize}
    \item ScienceQA, which includes science-related questions and images, requiring reasoning over visual and textual information~\citep{lu2022learn}. 
    \item  Seismic, a multimodal dataset related to seismic events capturing geographical images, and textual description of events.
    This work focuses on interpreting geological images with an open-ended question/answering dataset created manually by subject matter experts.
    \item VQA-Scene, a dataset for Visual Question Answering (VQA) centered on scene understanding~\citep{zhang2016yin}. We use a random sample of 1000 instances from the validation split of the VQA-Scene dataset
\end{itemize}

We evaluate models on these datasets chosen for their diversity, which is important for comparing different LLM backbones. 
In particular, ScienceQA emphasizes memorization‑driven language reasoning~\citep{yue2024mmmumassivemultidisciplinemultimodal,horawalavithana2023scitune}, while Seismic and VQA‑Scene minimize vision and language priors, respectively~\citep{zhang2016yin}. 
Please see Appendix~\ref{sec:vqa_scene_performance},~\ref{sec:sqa_performance} and~\ref{sec:seismic_performance} for sample queries.


\subsection{Evaluation Metrics}
\label{sec:eval_metrics}
To assess the model’s performance, we use standard metrics for each downstream task. 
For ScienceQA, we measure the accuracy for predicting the right choice. 
We use 2017 test instances from the ScienceQA dataset where the question, image and context are included in the input prompt.
For the Seismic dataset, we use the standard text-evaluation metrics (e.g., BLEU, ROUGE) for text outputs excluding the predictions for the latitude and longitude coordinates.
We calculate Haversine distance for the latitude and longitude values predicted by the model after parsing from the text output.
We normalize the distance to a score between 0 and 1 using a linear decay. 
Any distance exceeding a pre-defined maximum of 5,000 km is assigned a score of 0. 
For all other values, the score is calculated by the ratio of the distance to the maximum threshold, where a distance of 0 km yields a score of 1.
We used the VQA evaluation metric~\citep{antol2015vqa} which is computed based on the number of human subjects who provided the predicted answer (scaled by 3 and
clipped at 1).
For binary yes/no questions in the VQA-Scene dataset, we used the 10 human answers as provided in the dataset~\citep{antol2015vqa}.

\section{Empirical Analysis}
In this section, we analyze the downstream VLM performance across the three datasets, ScienceQA, VQA-Scene, and Seismic.
We present results when varying the LLM backbone from LLAMA-1 through LLAMA-2 to LLAMA-3.

\begin{figure*}[!t]
    \centering
    \begin{subfigure}[b]{0.32\textwidth}
        \centering
        \includegraphics[width=\textwidth]{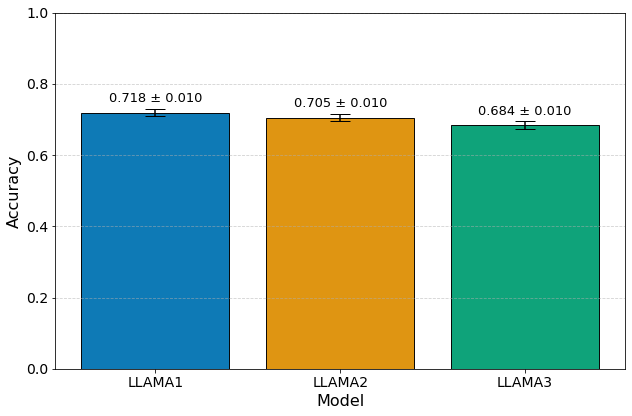}
        \caption{ScienceQA Accuracy}
        \label{fig:LLAMA_scienceqa_accuracy}
    \end{subfigure}
    \hfill
    \begin{subfigure}[b]{0.32\textwidth}
        \centering
        \includegraphics[width=\textwidth]{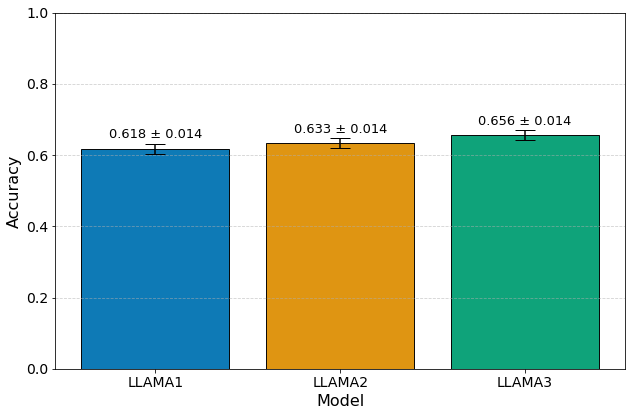}
        \caption{VQA Accuracy}
        \label{fig:LLAMA_vqa_accuracy}
    \end{subfigure}
    \hfill
    \begin{subfigure}[b]{0.32\textwidth}
        \centering
        \includegraphics[width=\textwidth]{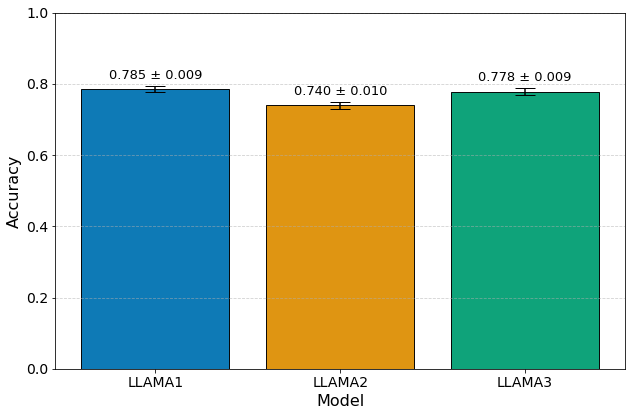}
        \caption{Seismic Accuracy}
        \label{fig:LLAMA_seismic_accuracy}
    \end{subfigure}
    \hfill
    \begin{subfigure}[b]{0.32\textwidth}
        \centering
        \includegraphics[width=\textwidth]{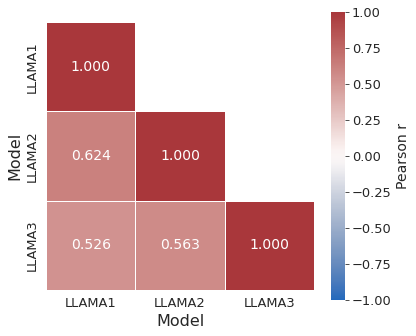}
        \caption{ScienceQA Instance-level Performance Correlation}
        \label{fig:LLAMA_scienceqa_corr}
    \end{subfigure} 
    \hfill
    \begin{subfigure}[b]{0.32\textwidth}
        \centering
        \includegraphics[width=\textwidth]{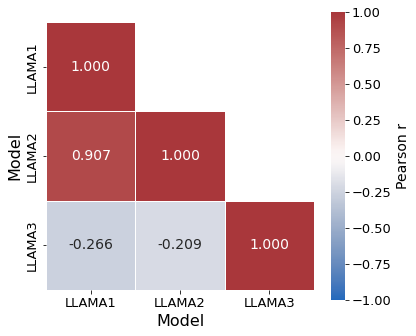}
        \caption{VQA Test Instance-level Performance Correlation}
        \label{fig:LLAMA_vqa_corr}
    \end{subfigure}
    \hfill
    \begin{subfigure}[b]{0.32\textwidth}
        \centering
        \includegraphics[width=\textwidth]{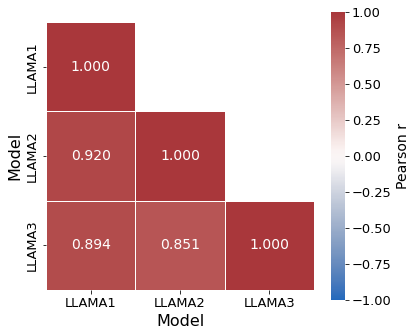}
        \caption{Seismic Instance-level Performance Correlation}
        \label{fig:LLAMA_seismic_corr}
    \end{subfigure}
    \caption{VLM performance with evolving pretrained LLM backbones across various domains. }
    \label{fig:LLAMA_runs}
\end{figure*}

\subsection{Overall Performance Across LLM Backbones}
We compare mean accuracy across the three datasets for each LLAMA backbone (Figures~\ref{fig:LLAMA_scienceqa_accuracy} to~\ref{fig:LLAMA_seismic_accuracy}).
There is no consistent pattern where a newer LLM leads to a better VLM.
For example, ScienceQA accuracy drops from 71.8\% (LLAMA-1) to 68.4\% (LLAMA-3).
VQA-Scene accuracy increases from 61.8\% to 65.6\% over LLAMA-1 to LLAMA-3.
Seismic text-output accuracy remains mixed, where the accuracy decreases from 78.5\% to 74.0\% and back to 77.8\% over the base LLAMA backbones.

Since the vision encoder, and training data are held constant, these differences reflect how each task interacts with the evolving capabilities of the LLM backbone.
We found different categories of VLM tasks have different impacts from LLM upgrades, as they will not always improve performance.

\begin{tcolorbox}[
    colback=cyan!5!white,
    colframe=cyan!60!black,
    title=\textbf{Takeaway 1: Newer LLMs do not uniformly improve VLM performance.},
    fonttitle=\bfseries,
    boxrule=0.8pt, arc=3pt,
    left=6pt, right=6pt, top=6pt, bottom=6pt
]
Under controlled conditions, newer LLMs do not uniformly improve VLM performance. Accuracy increases on VQA-Scene (+3.8\%), decreases on ScienceQA ($-$3.4\%), and stays comparable on Seismic. The performance trend depends on the downstream task.
\end{tcolorbox}

\subsection{Same Modality, Different Outcomes on Reasoning Tasks}

Both ScienceQA and VQA-Scene require reasoning over visual and textual information, yet they exhibit opposite performance trend to LLM evolution.
On ScienceQA, accuracy drops by 3.4 percentage points from LLAMA-1 to LLAMA-3.
On VQA-Scene, accuracy improves by 3.8 percentage points over the same LLM backbones.
It is the only dataset where we observe that better LLM backbones result better VLM performance.

Several factors may explain the performance drop on ScienceQA.
ScienceQA training examples follows a structured format where a \texttt{LECTURE} block provides background knowledge, followed by a solution that applies this knowledge to arrive at the answer.
At test time, the model is expected to recall the relevant \texttt{LECTURE} content and generate a solution conditioned on it~\citep{liu2023llava}.
This setup largely tests memorization-driven language reasoning, the model's ability to store and reproduce structured text patterns from training rather than genuine visual understanding of the input image~\citep{horawalavithana2023scitune,liu2024llavanext}.
Due to these characteristics, ScienceQA seems highly amenable to SFT, given it leads to strong in-distribution memorization~\citep{chu2025sft}. 
Once the VLM learns effective image-to-text projection and reasoning patterns, further LLM improvements (better pre-training data, longer context, or architectural tweaks in LLAMA-3) yield diminishing returns.
It is also possible that stronger language priors in newer LLMs cause the model to rely more on textual shortcuts rather than attending to the visual input, a known failure mode where the language model overrides visual evidence~\citep{Qiaoetal2024}.
On the other hand, VQA-Scene requires grounded visual reasoning about everyday scenes, where improved language understanding translates more directly to better answers.

\begin{tcolorbox}[
    colback=cyan!5!white,
    colframe=cyan!60!black,
    title=\textbf{Takeaway 2: Not all VLM reasoning tasks benefit equally from LLM upgrades},
    fonttitle=\bfseries,
    boxrule=0.8pt, arc=3pt,
    left=6pt, right=6pt, top=6pt, bottom=6pt
]
ScienceQA and VQA-Scene both require reasoning over visual input, yet they respond oppositely to LLM upgrades. The nature of the reasoning, memorization-driven versus grounded visual determines whether a newer LLM helps or not. Stronger LLMs do not universally improve reasoning-dependent VLM tasks.
\end{tcolorbox}

\subsection{Newer LLM Backbones Solve Different VLM Problems, Not Just More Problems}
To understand whether newer LLMs solve the same instances or shift to different subsets, we compute instance-level accuracy correlations across LLAMA generations (Figures~\ref{fig:LLAMA_scienceqa_corr} to~\ref{fig:LLAMA_seismic_corr}).
On ScienceQA, correlations are moderate and drop across LLM backbones.
LLAMA-1 and LLAMA-2 correlates at $r = 0.624$, while LLAMA-1 and LLAMA-3 agree at only $r = 0.526$ (Figures~\ref{fig:LLAMA_scienceqa_corr}).
This means LLAMA-3 does not simply lose accuracy on a fixed set of questions.
It gets some questions right that LLAMA-1 misses, and misses others that LLAMA-1 gets right.
LLAMA-1 and LLAMA-2 are nearly interchangeable ($r = 0.907$), solving almost the same questions on VQA-Scene dataset (Figures~\ref{fig:LLAMA_vqa_corr}).
However, LLAMA-3 is \textit{negatively correlated} with both LLAMA-1 ($r = -0.266$) and LLAMA-2 ($r = -0.209$).
Questions that LLAMA-1 and LLAMA-2 answer correctly tend to be the ones LLAMA-3 gets wrong, and vice versa.
The higher overall accuracy of LLAMA-3 does not come from solving the same questions plus a few more, but it comes from solving a largely \textit{different} set of questions.
Because VQA-Scene is designed so that language priors are controlled and vision plays an essential role~\citep{zhang2016yin}, this behavioral shift suggests that LLAMA-3 engages with visual information in a different way than its predecessors (see Appendix~\ref{sec:vqa_scene_performance} for more examples).
On Seismic, all correlations remain high ($r > 0.85$), confirming that the three models produce nearly identical results at the instance level.

\begin{tcolorbox}[
    colback=cyan!5!white,
    colframe=cyan!60!black,
    title=\textbf{Takeaway 3: Overall accuracy hides instance-level model behavioral shifts},
    fonttitle=\bfseries,
    boxrule=0.8pt, arc=3pt,
    left=6pt, right=6pt, top=6pt, bottom=6pt
]
On VQA-Scene, LLAMA-3 achieves higher accuracy than LLAMA-1/2 but is negatively correlated at the instance level. It succeeds on a largely different set of questions. Reporting only overall accuracy can hide substantial shifts in what the model actually solves.
\end{tcolorbox}

\subsection{When Vision Is the Bottleneck, the LLM Backbone Has Minimal VLM Performance Advantage}
On the Seismic dataset, we evaluate two types of output: text-based descriptions of geological features and numerical predictions of latitude and longitude coordinates.
As shown in Figure~\ref{fig:LLAMA_seismic_accuracy}, the text-based metrics show no systematic trend across LLM backbones.
Accuracy is 78.5\% for LLAMA-1, 74.0\% for LLAMA-2, and 77.8\% for LLAMA-3.
Instance-level correlations are consistently high ($r > 0.85$), meaning all three models succeed and fail on nearly the same instances (Figure~\ref{fig:LLAMA_seismic_corr}).

Interpreting geological images depends primarily on the visual encoder's ability to extract domain-specific features.
In this setup, LLM backbone generates texts primarily based on whatever the visual encoder provides as geological features.
Since CLIP is held constant across experiments, the vision perception challenges stay the same, and LLM improvements have little room to make a difference.
This finding is consistent with prior work calling for advances in vision encoders to improve VLM performance on perception-heavy tasks~\citep{tong2024cambrian}.

\begin{tcolorbox}[
    colback=cyan!5!white,
    colframe=cyan!60!black,
    title=\textbf{Takeaway 4: LLM backbones have minimal performance advantage over perception-heavy VLM tasks},
    fonttitle=\bfseries,
    boxrule=0.8pt, arc=3pt,
    left=6pt, right=6pt, top=6pt, bottom=6pt
]
On the Seismic text-based evaluation, all three LLAMA backbones solve nearly identical instances ($r > 0.85$). When the bottleneck is at the visual feature extraction, upgrading the LLM backbone offers minimal returns. For such tasks, improving the vision encoder is more likely to yield meaningful gains.
\end{tcolorbox}

\subsection{Newer LLM Backbones Exhibit New VLM Capabilities That Older LLMs Lack}



\begin{figure}[!t]
  \centering
  \begin{minipage}{0.48\textwidth}
    \centering
    \includegraphics[width=\textwidth]{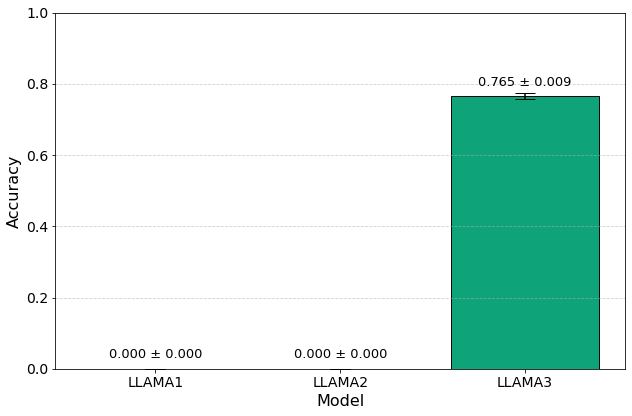}
        \caption{Accuracy on predicting the Latitude/Longitude coordinate values in the Seismic dataset.}
\label{fig:LLAMA_seismic_accuracy_latlong}
  \end{minipage}
  \hfill 
  \begin{minipage}{0.45\textwidth}
    \centering
    \includegraphics[width=\textwidth]{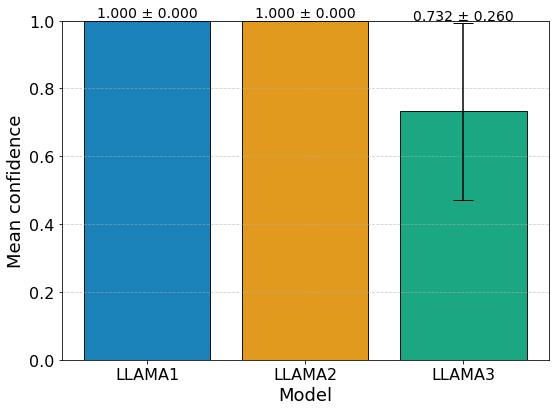}
        \caption{Model Confidence in the VQA Scene Dataset}
    \label{fig:LLAMA_vqa_confidence}
  \end{minipage}
\end{figure}

While the text-based Seismic results suggest that LLM backbone evolution has little effect on the text-output performance, the coordinate prediction task demonstrate a different performance pattern.
In this task, we evaluate the model's ability to predict latitude and longitude values from geological images using the Haversine-based scoring described in Section~\ref{sec:eval_metrics}.
LLAMA-1 and LLAMA-2 both score exactly 0.0\% on this metric—they are completely unable to predict valid coordinate predictions, instead predict the text listing the city and state names (see Appendix~\ref{sec:seismic_performance}).
LLAMA-3, by contrast, records 76.5\% accuracy on the same instances (Figure~\ref{fig:LLAMA_seismic_accuracy_latlong}).

There are several possible explanations for this performance increase.
LLAMA-3 was pre-trained on a substantially larger and more diverse corpus, which likely includes more geospatial data, coordinate formats, and structured numerical content.
Its larger vocabulary (128K tokens) may also represent numerical values more effectively.
Additionally, the improved instruction-following capabilities of LLAMA-3 may allow it to parse and respond to the structured output format required for coordinate prediction, a task that demands accurate geospatial formatting rather than open-ended generation.

This result re-emphasizes the perception-based conclusion from the previous subsection.
The text-based description of geological features is indeed bottlenecked by the visual encoder where all three LLMs perform similarly.
But the coordinate prediction task also needs another skill where the LLM needs to predict formatted numbers that match specific locations on the map.

\begin{tcolorbox}[
    colback=cyan!5!white,
    colframe=cyan!60!black,
    title=\textbf{Takeaway 5: Newer LLM backbones introduce new VLM capabilities},
    fonttitle=\bfseries,
    boxrule=0.8pt, arc=3pt,
    left=6pt, right=6pt, top=6pt, bottom=6pt
]
On the Seismic coordinate prediction task, LLAMA-1 and LLAMA-2 score 0.0\% while LLAMA-3 scores 76.5\%. VLM capabilities which require structured numerical output or domain-specific knowledge may emerge only with sufficiently advanced LLM backbones.
\end{tcolorbox}

\subsection{VLM Confidence and Internal Contextual Representations}


\begin{figure*}[!t]
    \centering
    \begin{subfigure}[b]{0.53\textwidth}
        \centering
        \includegraphics[width=\textwidth]{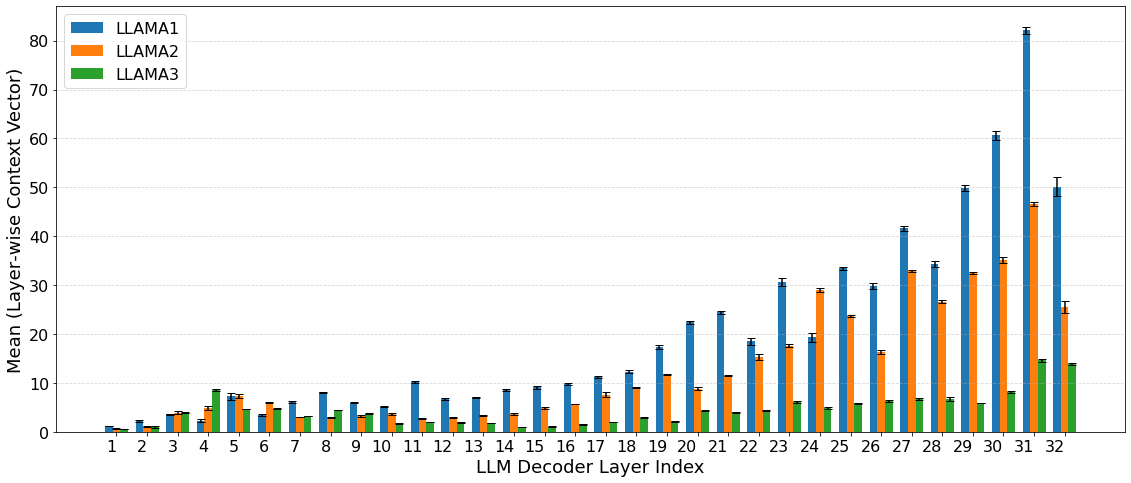}
        \caption{Mean layer-wise context vector}
        \label{fig:LLAMA_vqa_layer_mean}
    \end{subfigure}
    \hfill
    \begin{subfigure}[b]{0.45\textwidth}
        \centering
        \includegraphics[width=\textwidth]{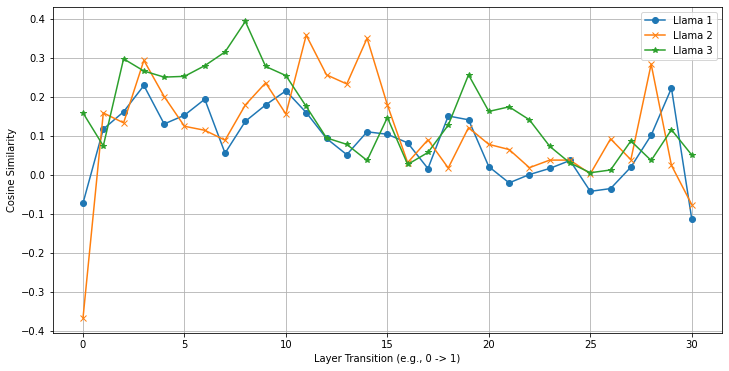}
        \caption{Cosine similarity across layer-wise context vector}
        \label{fig:LLAMA_vqa_layer_transition}
    \end{subfigure}
    \caption{Analysis on layer-wise context vectors of the VLMs trained and tested on VQA Scene Dataset. Layer k's context vector encodes the model state about what token to generate next, given the question and image.}
\end{figure*}

The instance-level analysis in the previous section revealed that LLAMA-3 solves a largely different set of VQA-Scene questions compared to LLAMA-1 and LLAMA-2.
To understand the mechanisms behind this behavioral shift, we conduct additional analysis on the VQA-Scene predictions.
We examine two aspects of the model's internals: the token-level confidence of generated answers, and the layer-wise representations within the decoder.

\subsubsection{LLAMA-3 based VLM Is Less Confident but More Accurate}
We extract log-probabilities of the generated answer tokens and convert them to confidence scores using $c = \exp(\mathsf{log\_prob})$.
A confidence of 1.0 means the model assigns near-certain probability to its predicted answer.
Figure~\ref{fig:LLAMA_vqa_confidence} shows the mean confidence scores across VQA-Scene instances for each model.

LLAMA-1 and LLAMA-2 both produce confidence scores of 1.0, meaning they assign near-certain probability to their answers on every instance.
LLAMA-3, by contrast, has a mean confidence of 0.73, substantially lower score.
Yet LLAMA-3 achieves the highest accuracy on this dataset (65.6\% vs.\ 61.8\% and 63.3\%).
The model that is least sure of its answers is the one that gets the most answers right.

LLAMA-1 and LLAMA-2 are overconfident where they assign maximum probability to their outputs regardless of whether they are correct.
LLAMA-3 distributes probability mass more carefully across possible answers, resulting in lower per-token confidence but better overall performance.
It also helps explain the negative instance-level correlation: if LLAMA-1 and LLAMA-2 always commit fully to one answer while LLAMA-3 hedges differently, the models will naturally arrive at different answers on many instances even when the underlying visual information is the same.

\begin{tcolorbox}[
    colback=cyan!5!white,
    colframe=cyan!60!black,
    title=\textbf{Takeaway 7: Higher accuracy does not require higher confidence},
    fonttitle=\bfseries,
    boxrule=0.8pt, arc=3pt,
    left=6pt, right=6pt, top=6pt, bottom=6pt
]
On VQA-Scene, LLAMA-3 achieves the highest accuracy while being the least confident (0.73 vs.\ 1.0 for LLAMA-1/2). Older LLM backbones produce overconfident VLMs that commit fully to answers regardless of correctness. Newer LLMs yield better-calibrated predictions, which partly explains why they solve different instances.
\end{tcolorbox}

\subsubsection{Internal Representations Diverge Across LLM Generations}

To further understand how the three models differ, we examine the internal representations produced during answer generation.
Specifically, we record the context layer output, the final representation vector after self-attention at each of the 32 decoder layers for the generated answer token.
Each context vector has dimensionality 4096.
We analyze two properties: the magnitude of these representations across layers, and the similarity of representations across models at each layer.

\paragraph{Contextual representation magnitude.}
Figure~\ref{fig:LLAMA_vqa_layer_mean} shows the mean magnitude of the context vectors across layers for each model.
The three models show qualitatively different activation profiles.
LLAMA-1 produces the largest activations, with magnitudes growing steeply through deeper layers and peaking at 82.2 in layer 31 before dropping to 50.2 in the final layer.
LLAMA-2 follows a similar growth pattern but at roughly half the scale, peaking at 46.5.
LLAMA-3's magnitudes remain much smaller throughout, growing slowly and reaching only 14.6 at layer 31.

This progression from large growing activations in LLAMA-1 to small activations in LLAMA-3 may reflect the architectural and training improvements across models.
LLAMA-3 uses grouped query attention and was trained with improved normalization strategies, both of which contribute to more stable and compact internal representations.
Three LLM backbones process visual and textual information through very different internal dynamics, even when given identical inputs.
A model with small activations is unlikely to produce the same outputs as one with large, variable activations consistent with the different instance-level behavior we observed earlier.

\paragraph{Layer-to-layer contextual representation transition.}
Beyond magnitude, we measure how each model transforms its context representations from one layer to the next.
For each model, we compute the cosine similarity between the context vectors of consecutive layers, layer $l$ and layer $l{+}1$ across all 32 decoder layers (Figure~\ref{fig:LLAMA_vqa_layer_transition}).
High similarity between adjacent layers means the representation changes gradually; low or negative similarity means the representation has an significant transformation at that transition.

The three models exhibit distinct layer-to-layer transition profiles.
LLAMA-1 shows moderate similarities in the early and middle layers (roughly 0.1--0.2), but in the deeper layers from around layer 20 onward the similarities drop sharply, reaching near-zero and negative values.
This means the representation has significant transformations in the later layers where the final answer is being formed.
LLAMA-2 follows a broadly similar pattern where moderate-to-high similarities in the middle layers (peaking at 0.36 around layer 11), followed by a drop to near-zero in the deeper layers, with a brief spike at layer 28 before turning negative again.
This aligns with their high instance-level correlation ($r = 0.907$) (see Figure~\ref{fig:LLAMA_vqa_corr}) as they process information through a similar internal pathway and arrive at the same answers.

LLAMA-3's layer-to-layer similarities are higher and more stable in the early layers (0.15--0.39), and while they drops gradually in the deeper layers.
LLAMA-3 has no negative layer transitions at all.
This aligns with the negative instance-level correlation ($r = -0.266$ and $r = -0.209$) (see Figure~\ref{fig:LLAMA_vqa_corr})  as the model predicts its answers through a qualitatively different internal process.

This means LLAMA-3 builds its answer representation through incremental refinements at every layer, while LLAMA-1 and LLAMA-2 have significant contextual representational shifts in the deeper layers.
In older LLM backbones, the later decoder layers appear to significantly reorganize the contextual representation before producing the final answer, a process that may be less stable and more prone to committing to incorrect outputs with high confidence.
LLAMA-3's processing pipeline is consistent with both its better calibration (it does not overcommit) and its different instance-level behavior (it arrives at answers through a different internal pathway).
This has implications for how we evaluate VLM progresss, two models with similar aggregate accuracy can have very different internal mechanisms, and a model with higher accuracy may achieve it through an entirely different route than its predecessor.

\begin{tcolorbox}[
    colback=cyan!5!white,
    colframe=cyan!60!black,
    title=\textbf{Takeaway 8: VLMs with newer LLM backbones process information through a different internal pathway},
    fonttitle=\bfseries,
    boxrule=0.8pt, arc=3pt,
    left=6pt, right=6pt, top=6pt, bottom=6pt
]
LLAMA-3 based VLMs predict the answer through incremental refinements over the layers. LLAMA-1 and LLAMA-2 rely on large activations and significant contextual shifts in the deeper layers. These internal differences may provide a mechanistic explanation for why the models solve different instances despite receiving identical visual and textual inputs.
\end{tcolorbox}

\section{Conclusion}
In this work, we study whether newer LLM backbones lead to better VLM downstream task performance.
To answer this question, we designed a controlled study where we change the LLM backbone from LLaMA-1 through LLaMA-2 to LLaMA-3 while the vision encoder, training data, and codebase remain the same when training a VLM.
We evaluated the resulting VLMs on three downstream tasks spanning different domains.

Our main takeaways are as follows.
First, a newer LLM backbone does not consistently produce a better VLM, but the performance depends entirely on the downstream task.
Second, newer LLM backbones do not just solve the same questions better.
They solve \textit{different} questions, and aggregate accuracy alone can hide these behavioral shifts.
Third, we noticed that newer LLM backbones process information through a qualitatively different pathway with better-calibrated confidence.
Fourth, certain VLM capabilities can emerge in newer LLM generations rather than improving gradually, particularly those requiring structured numerical output or domain-specific knowledge.
Finally, when a task is bottlenecked by visual perception, the LLM backbone is largely interchangeable, and improving the vision encoder is more likely to yield meaningful gains.

\section*{Acknowledgements}
This work was supported by the NNSA Office of Defense Nuclear Nonproliferation Research and Development, U.S. Department of Energy, and Pacific Northwest National Laboratory, which is operated by Battelle Memorial Institute for the U.S. Department of Energy under Contract DE-AC05–76RLO1830. This article has been cleared by PNNL for public release as PNNL-SA-221530.




\bibliography{bib-v1,bib-v2}
\bibliographystyle{colm2026_conference}

\appendix
\section{Appendix: Hyper-parameters}
Table~\ref{tab:hyperparams} summarizes the key hyperparameters used in our multimodal NeVA fine‑tuning experiments. 
To improve clarity and reproducibility, we report only the parameters that differ between the public NeVA default configuration~\citep{NVIDIANeMo_neva_config_yaml_v2_2_1} and our three model variants: Llama~3.1 (8B), Llama~2 (7B), and the 
Llama~1 (7B). Parameters identical across all settings—such as mixed‑precision (BF16), 
Transformer Engine, Megatron‑AMP O2, parallelism configuration (TP{=}1, PP{=}1), optimizer 
(fused Adam), and checkpointing behavior are intentionally omitted to avoid redundancy.
We retain only the best-performing checkpoint according to the validation metric used during training.
The original LLAMA model checkpoints were converted into the \texttt{.nemo} format prior to fine-tuning in order to ensure compatibility with the NeMo Megatron training pipeline.
All experiments were conducted using NVIDIA A100 GPUs, with a total of 8 GPUs available for training.

\begin{table}[htbp]
\centering
\scriptsize
\caption{Hyperparameters across NeVA default vs. three LLAMA profiles.
Identical values (BF16, TP=1, PP=1, fused\_adam, etc.) are omitted.}
\label{tab:hyperparams}
\begin{tabular}{lllll}
\hline
\textbf{Parameter} & \textbf{NeVA Default (Image)} & \textbf{LLAMA 3.1 Instruct (8B)} & \textbf{LLAMA 2 (7B)} & \textbf{LLAMA 1 (7B)} \\
\hline
\multicolumn{5}{l}{\textbf{Model Scale / Transformer Architecture}} \\
encoder\_seq\_length    & 4096\textsuperscript{*} & 8192 & 4096 & 2048 \\
num\_layers             & 24\textsuperscript{*}    & 32   & 32   & 32   \\
hidden\_size            & 2048\textsuperscript{*}  & 4096 & 4096 & 4096 \\
ffn\_hidden\_size        & 5440\textsuperscript{*}  & 14336 & 11008 & 11008 \\
num\_attention\_heads    & 16\textsuperscript{*}    & 32   & 32   & 32   \\
\hline
\multicolumn{5}{l}{\textbf{Batching}} \\
micro\_batch\_size       & 16\textsuperscript{*}            & 2                       & 2                       & 2 \\
global\_batch\_size      & 128\textsuperscript{*}           & 16                      & 16                      & 16 \\
\hline
\end{tabular}

\vspace{2pt}
{\footnotesize \textsuperscript{*}\,Default values summarized from the public NeVA v2.2.1 config.}
\end{table}
\section{Appendix: VLM Pre-training Performance}
\label{sec:vlm_scicap}
In the first stage of the training, referred to as the warm-up or alignment stage, the objective is to align the visual encoder with the LLM decoder.
Table~\ref{tab:scicap_performance} summarizes the performance of the models resulted from this stage for each model variant on the SciCap test set.
Across all three model variants, the BLEU and ROUGE metrics on the SciCap test set are highly comparable.
This consistency demonstrates that the alignment stage produce stable initial VLM checkpoints regardless of the underlying LLM backbone. 

\begin{table}[!t]
\centering
\caption{VLM performance of image captioning on the SciCap test set with different LLM backbones. 
Metrics are reported as mean ($\mu$) and standard deviation ($\sigma$) for the best validated checkpoint of each model.}
\label{tab:scicap_performance}
\begin{tabular}{lcc}
\hline
\textbf{Model} & \textbf{BLEU ($\mu \pm \sigma$)} & \textbf{ROUGE ($\mu \pm \sigma$)} \\
\hline
LLAMA 1 & $0.02046 \pm 0.01791$ & $0.06444 \pm 0.04520$ \\
LLAMA 2 & $0.02178 \pm 0.01786$ & $0.06860 \pm 0.04723$ \\
LLAMA 3 & $0.01967 \pm 0.01800$ & $0.06382 \pm 0.04764$ \\
\hline
\end{tabular}
\caption{Example predictions on ScienceQA Dataset. In the ground truth answer choices, the bold answer is the ground truth. For brevity, we omit the LECTURE block in the prediction text.}
\end{table}
\section{VQA-Scene Test Examples}
\label{sec:vqa_scene_performance}

Table~\ref{tab:vqa_scene_performance} shows random test examples and their predictions across the LLAMA models.
Note that, each question is annotated with ten human answers, denoted 
$\mathcal{A}=\{a_i\}_{i=1}^{10}$. Let $\hat{a}$ be the model's predicted answer.
We first count how many human annotations exactly match the prediction:
\[
\text{matching\_count} \;=\; \bigl|\{\, i \in \{1,\dots,10\} \;:\; a_i = \hat{a} \,\}\bigr|.
\]
The VQA accuracy for this question is then computed as
\[
\text{score} \;=\; \min\!\left(\frac{\text{matching\_count}}{3},\, 1.0\right).
\]
\noindent
In words, if at least three out of ten humans provided the same answer as the model, 
the prediction is considered fully correct ($\text{score}=1$); otherwise, 
partial credit is awarded in increments of $\frac{1}{3}$ (e.g., one match $\frac{1}{3}$, 
two matches $\frac{2}{3}$).

Consider the question: \emph{“Is the table sturdy?”} applied to the scene shown in the image (see fourth row in Table~\ref{tab:vqa_scene_performance}). 
The ground‐truth annotations consist of ten human answers. 
In this example, seven annotators responded “yes” and three responded “no.” 
Let $\hat{a}$ be the model’s predicted answer. If the model also predicts “no,” 
the number of matching human responses is
\[
\text{matching\_count} = 3.
\]
Following the standard VQA accuracy protocol, the item score is computed as
\[
\text{score} = \min\!\left(\frac{\text{matching\_count}}{3},\, 1.0\right)
             = \min\!\left(\frac{3}{3},\, 1.0\right)
             = 1.0.
\]
Because at least three annotators provided the same answer as the model, the prediction 
is considered fully correct under the original VQA metric.
This illustrates a known caveat of the VQA metric, whenever at least three annotators agree on an answer, that answer is treated as fully correct. 
The metric intentionally behaves this way to reflect the 
inherent ambiguity and abstractness of certain visual questions, where multiple human answers can be considered valid and the level of agreement serves as a proxy for collective confidence.

\begin{table}[!t]
\centering
\scriptsize
\caption{Example predictions on VQA-Scene Dataset. We list the ground truth answer as what recorded by the majority of human annotators. We used the same query on all 3 model generations.}
\label{tab:vqa_scene_performance}

\begin{tabular}{p{2.4cm} p{3.2cm} p{1.6cm} p{1.6cm} p{1.6cm}}
\toprule
\textbf{Image} & \textbf{Query/Ground Truth Answer} & \textbf{LLAMA-1 Answer} & \textbf{LLAMA-2 Answer} & \textbf{LLAMA-3 Answer} \\
\midrule
\raisebox{-0.5\height}{\includegraphics[width=2.4cm]{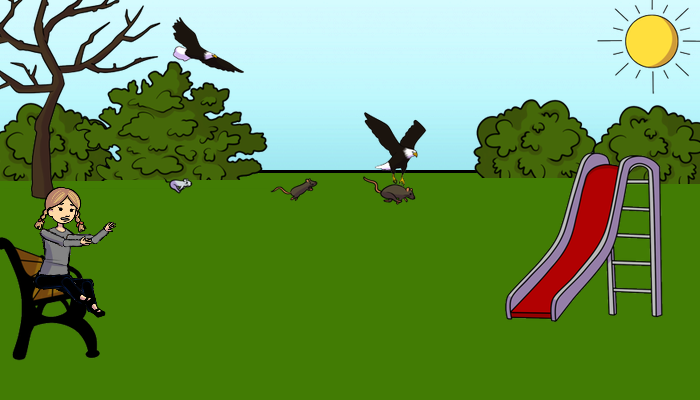}} &
Is the girl going down the slide?/\textbf{no} &
yes &
yes &
\textbf{no} \\
\raisebox{-0.5\height}{\includegraphics[width=2.4cm]{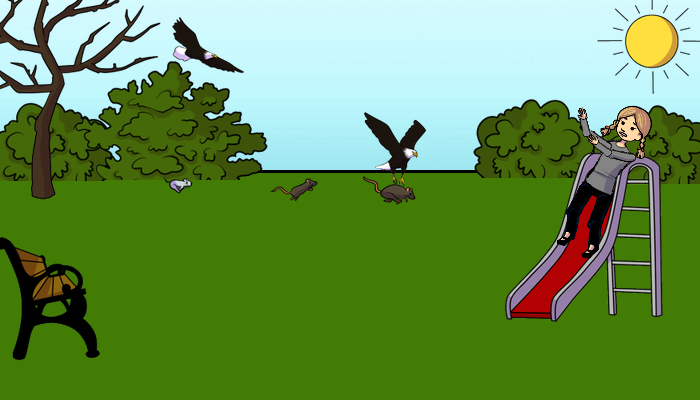}} &
Is the girl going down the slide?/\textbf{yes} &
\textbf{yes} &
\textbf{yes} &
no \\
\raisebox{-0.5\height}{\includegraphics[width=2.4cm]{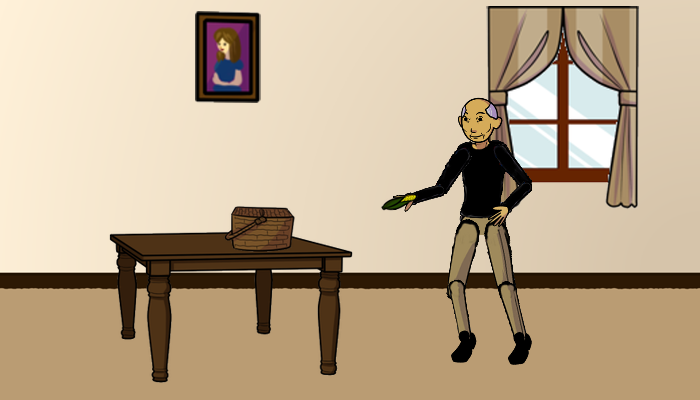}} &
Is the table sturdy?/\textbf{yes} &
\textbf{yes} &
\textbf{yes} &
no \\
\raisebox{-0.5\height}{\includegraphics[width=2.4cm]{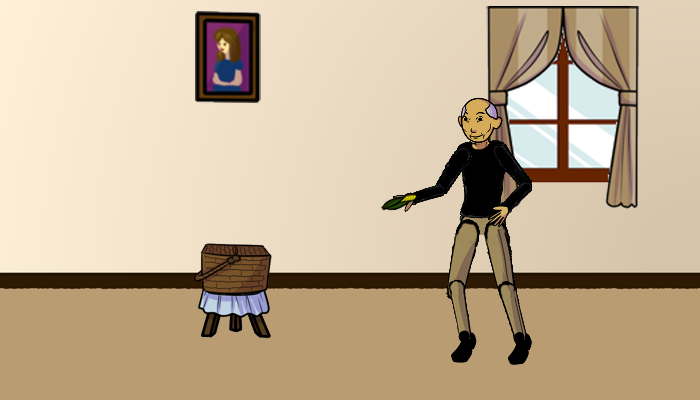}} &
Is the table sturdy?/\textbf{yes} &
no &
no &
no \\
\raisebox{-0.5\height}{\includegraphics[width=2.4cm]{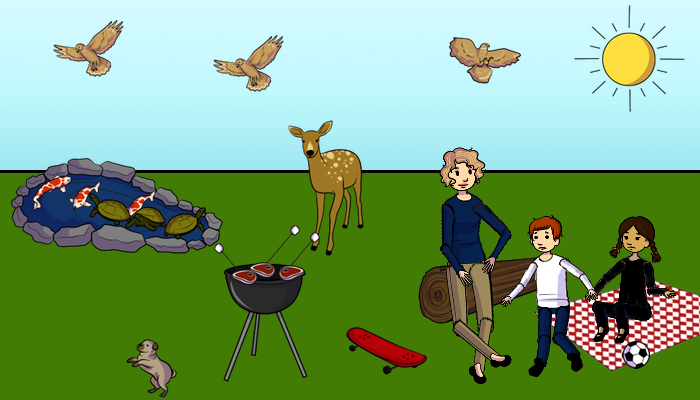}} &
Are the birds going to swoop down and take the meat from the grill?/\textbf{no} &
yes &
yes &
\textbf{no} \\
\raisebox{-0.5\height}{\includegraphics[width=2.4cm]{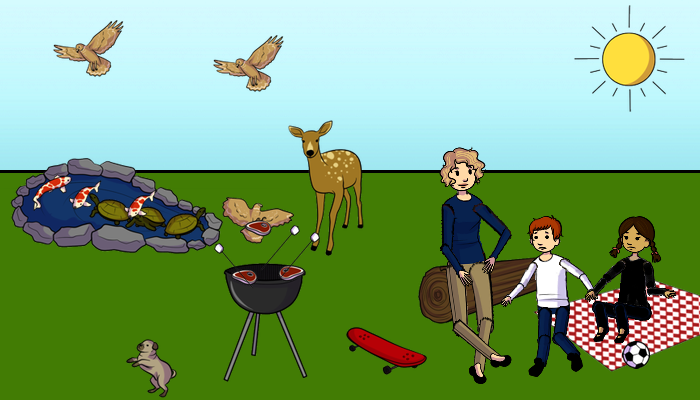}} &
Are the birds going to swoop down and take the meat from the grill?/\textbf{no}  &
yes &
yes &
yes \\
\raisebox{-0.5\height}{\includegraphics[width=2.4cm]{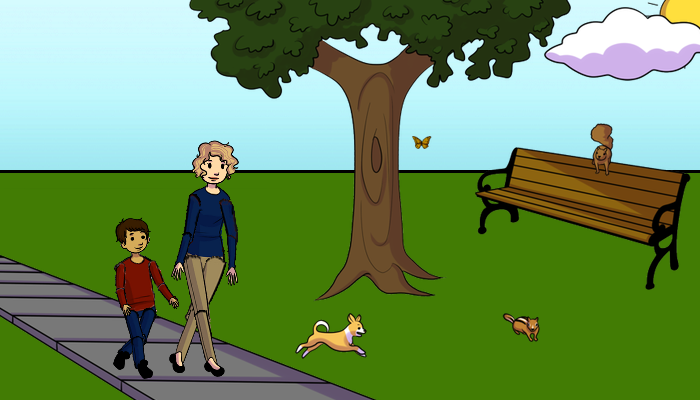}} &
Is the dog chasing after the squirrel?/\textbf{yes}  &
\textbf{yes}  &
\textbf{yes}  &
no \\
\raisebox{-0.5\height}{\includegraphics[width=2.4cm]{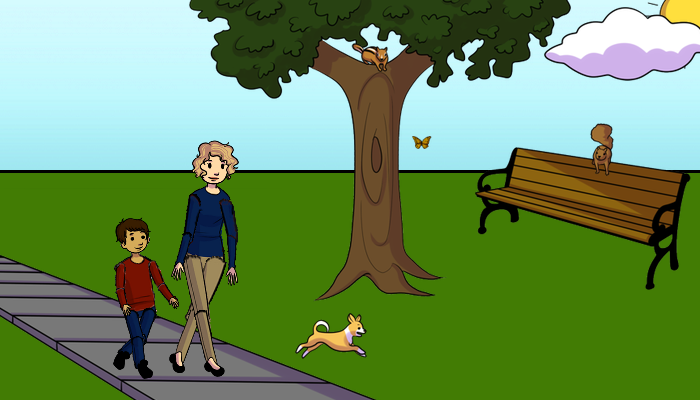}} &
Is the dog chasing after the squirrel?/\textbf{yes} &
\textbf{yes}  &
\textbf{yes}  &
\textbf{yes}  \\
\raisebox{-0.5\height}{\includegraphics[width=2.4cm]{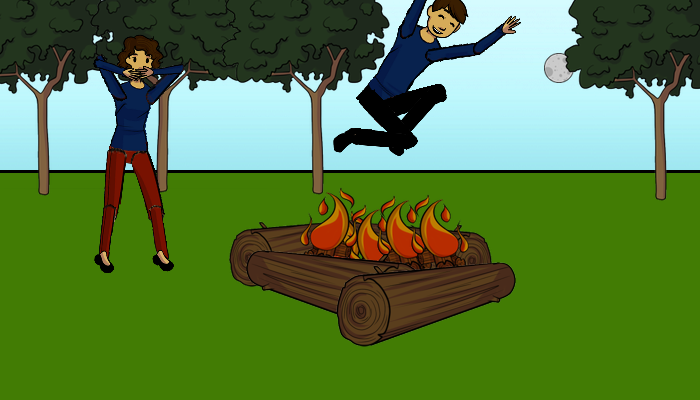}} &
Is it safe to jump over fire?/\textbf{no} &
yes  &
yes  &
\textbf{no}  \\
\raisebox{-0.5\height}{\includegraphics[width=2.4cm]{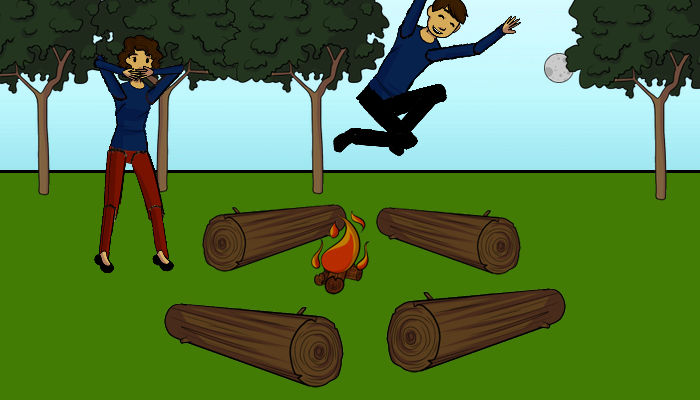}} &
Is it safe to jump over fire?/\textbf{no} &
yes  &
yes  &
yes  \\
\raisebox{-0.5\height}{\includegraphics[width=2.4cm]{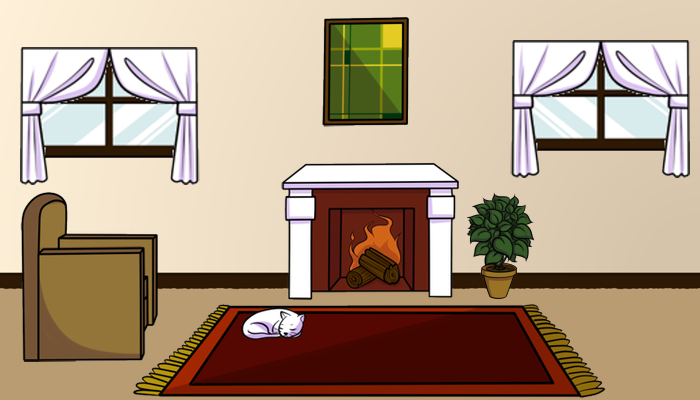}} &
Is the dog alone?/\textbf{yes} &
no  &
no  &
no  \\
\raisebox{-0.5\height}{\includegraphics[width=2.4cm]{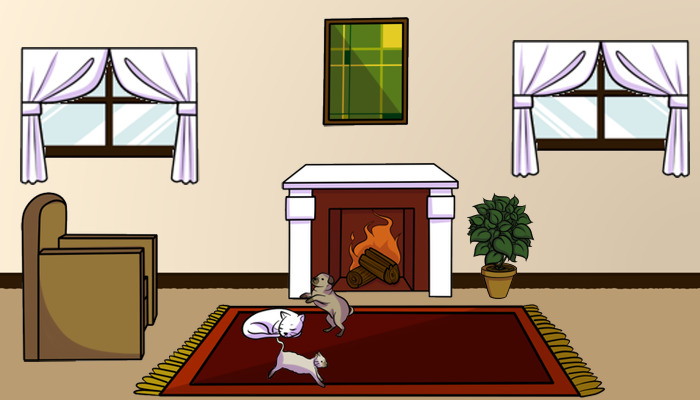}} &
Is the dog alone?/\textbf{no} &
yes  &
yes  &
\textbf{no}  \\
\bottomrule
\end{tabular}
\end{table}
\section{ScienceQA Test Examples}
\label{sec:sqa_performance}

Table~\ref{tab:sqa_performance} shows random test examples from the ScienceQA dataset.

\begin{table}[!t]
\centering
\scriptsize
\caption{Example predictions on ScienceQA Dataset. In the ground truth answer choices, the bold answer is the ground truth. For brevity, we omit the LECTURE block in the prediction text.}
\label{tab:sqa_performance}
\begin{tabular}{p{2.4cm} p{3.2cm} p{1.6cm} p{1.6cm} p{1.6cm}}
\toprule
\textbf{Image} & \textbf{Query/Ground Truth Answer} & \textbf{LLAMA-1 Answer} & \textbf{LLAMA-2 Answer} & \textbf{LLAMA-3 Answer} \\
\midrule
\raisebox{-0.5\height}{\includegraphics[width=2.4cm]{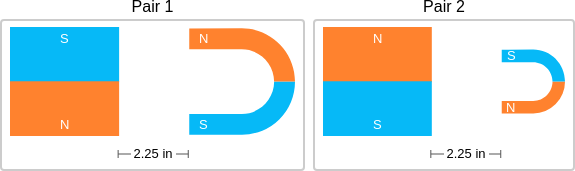}} &
Think about the magnetic force between the magnets in each pair. Which of the following statements is true?
Context: The images below show two pairs of magnets. The magnets in different pairs do not affect each other. All the magnets shown are made of the same material, but some of them are different sizes and shapes.
Options: (A) The magnitude of the magnetic force is the same in both pairs. (B) The magnitude of the magnetic force is smaller in Pair 1. \textbf{(C) The magnitude of the magnetic force is smaller in Pair 2.}
 &
SOLUTION: Magnet sizes affect the magnitude of the magnetic force. Imagine magnets that are the same shape and made of the same material. The smaller the magnets, the smaller the magnitude of the magnetic force between them.
Magnet A is the same size in both pairs. But Magnet B is smaller in Pair 1 than in Pair 2. So, the magnitude of the magnetic force is smaller in Pair 1 than in Pair 2.
\#\#\#
ANSWER: B. &
SOLUTION: Magnet sizes affect the magnitude of the magnetic force. Imagine magnets that are the same shape and made of the same material. The smaller the magnets, the smaller the magnitude of the magnetic force between them.
Magnet A is the same size in both pairs. But Magnet B is smaller in Pair 2 than in Pair 1. So, the magnitude of the magnetic force is smaller in Pair 2 than in Pair 1.
\#\#\#
\textbf{ANSWER: C.} &
SOLUTION: Magnet sizes affect the magnitude of the magnetic force. Imagine magnets that are the same shape and made of the same material. The smaller the magnets, the smaller the magnitude of the magnetic force between them.
Magnet A is the same size in both pairs. But Magnet B is smaller in Pair 1 than in Pair 2. So, the magnitude of the magnetic force is smaller in Pair 1 than in Pair 2.
\#\#\#
ANSWER: B. \\
\raisebox{-0.5\height}{\includegraphics[width=2.4cm]{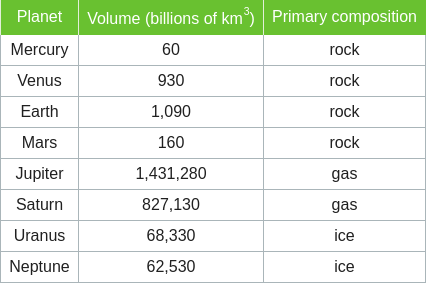}} &
Is the following statement about our solar system true or false?
The largest planet is made mainly of ice.
Context: Use the data to answer the question below.
Options: (A) true \textbf{(B) false}
&
SOLUTION: The table tells you that the largest planet is Jupiter and that Jupiter is made mainly of gas. So, the largest planet is made mainly of gas.
\#\#\#
ANSWER: A.
&
SOLUTION: To decide which planet is the largest, look at the volumes and compare the exponents. Jupiter's volume has an exponent of 15, which is the largest of all the planets. So, the largest planet is Jupiter.
Jupiter is made mainly of gas. So, the largest planet is made mainly of gas.
\#\#\#
\textbf{ANSWER: B.}
&
SOLUTION: The table tells you that the largest planet is Jupiter and that Jupiter is made mainly of gas. So, the largest planet is made mainly of gas.
\#\#\#
\textbf{ANSWER: B.}
\\
\bottomrule
\end{tabular}
\end{table}
\section{Seismic Test Examples}
\label{sec:seismic_performance}

Table~\ref{tab:seismic_performance} shows random test examples from the Seismic dataset.

\begin{table}[!t]
\centering
\scriptsize
\caption{Example predictions on Seismic Dataset.}
\label{tab:seismic_performance}
\begin{tabular}{p{2.4cm} p{3.2cm} p{1.6cm} p{1.6cm} p{1.6cm}}
\toprule
\textbf{Image} & \textbf{Query/Ground Truth Answer} & \textbf{LLAMA-1 Answer} & \textbf{LLAMA-2 Answer} & \textbf{LLAMA-3 Answer} \\
\midrule
\raisebox{-0.5\height}{\includegraphics[width=2.4cm]{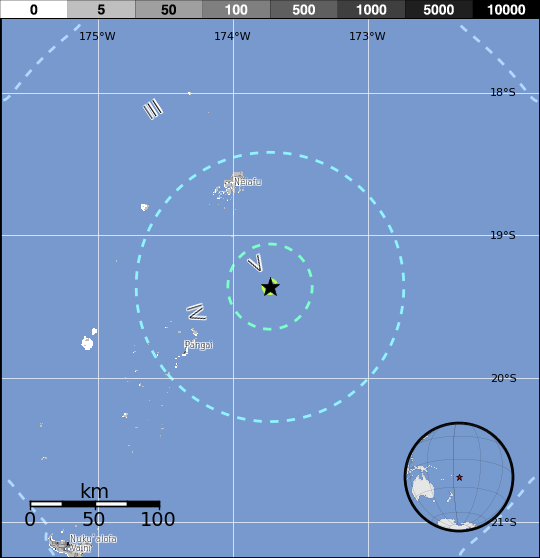}} & As per the exposure map, what is the latitude and longitude of the origin (star) of the epicenter in the seismic event "M 5.7 - 82 km NE of Pangai, Tonga"
Answer:
\textbf{Latitude: 19.361S Longitude: 173.728W}
&
Pangai, Neiafu, Nuku'alofa
&
Pangai, Nuku'alofa
&
\textbf{Latitude: 13.764S Longitude: 172.533W}
\\
\raisebox{-0.5\height}{\includegraphics[width=2.4cm]{images/seismic/usc000kn6g_1.png}} & As per the exposure map, determine whether the epicenter (star) of the seismic event "M 5.7 - 82 km NE of Pangai, Tonga" is on the land or water
Answer:
\textbf{water}
&
\textbf{water}
&
\textbf{water}
&
\textbf{water}
\\
\raisebox{-0.5\height}{\includegraphics[width=2.4cm]{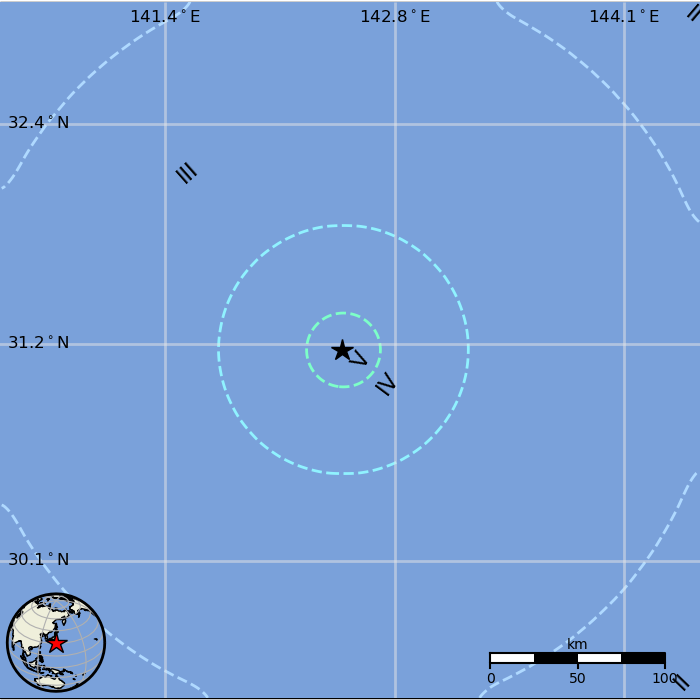}} & As per the exposure map, what is the latitude and longitude of the origin (star) of the epicenter in the seismic event "M 5.5 - Izu Islands, Japan region"
Answer:
\textbf{Latitude: 31.217N Longitude: 142.438E}
&
Tokyo, Nagoya
&
Tokyo, Nagoya
&
\textbf{Latitude: 32.827N Longitude: 130.754E}
\\
\raisebox{-0.5\height}{\includegraphics[width=2.4cm]{images/seismic/us6000g7tw_1.png}} & As per the exposure map, determine whether the epicenter (star) of the seismic event "M 5.5 - Izu Islands, Japan region" is on the land or water
Answer:
\textbf{water}
&
\textbf{water}
&
\textbf{water}
&
\textbf{water}
\\
\bottomrule
\end{tabular}
\end{table}

\end{document}